\documentclass[10pt,twocolumn,letterpaper]{article}

\usepackage[pagenumbers]{wacv} % arXiv version with page numbers, no review watermark/line numbers
\usepackage[T1]{fontenc}
\usepackage[utf8]{inputenc}
\usepackage{microtype}
\usepackage{graphicx}
\usepackage{cuted}
\graphicspath{{./}{prism-uploads/}}

\usepackage{booktabs}
\usepackage{multirow}
\usepackage{array}
\usepackage{xcolor}
\usepackage{amsmath}
\usepackage{amssymb}
\usepackage{tikz}
\usepackage{pgfplots}
\pgfplotsset{compat=1.18}

\newcommand{\safeincludegraphics}[2][]{%
  \IfFileExists{#2}{%
    \includegraphics[#1]{#2}%
  }{%
    \fbox{\parbox[c][0.28\textheight][c]{0.82\textwidth}{\centering Missing figure file:\\ \texttt{#2}}}%
  }%
}

\definecolor{wacvblue}{rgb}{0.21,0.49,0.74}
\usepackage[pagebackref,breaklinks,colorlinks,allcolors=wacvblue]{hyperref}

\def\wacvPaperID{*****} % *** Enter the WACV Paper ID here
\def\confName{WACV}
\def\confYear{2027}

\title{Do Video-LLMs Actually Watch? Diagnosing Character-Tracking\\
       Failures in Long-Form Video}

\author{Mohammad Al-Ratrout \and Shayla Sharmin \and Aditya Raikwar \and Roghayeh Leila Barmaki}

\begin{document}
\maketitle
\vspace{-2.0em}
% ---- Teaser (Figure 1): first-page full-width strip ----
\begin{strip}
\vspace{-6.2em}
\begin{center}
  \includegraphics[width=0.9\textwidth]{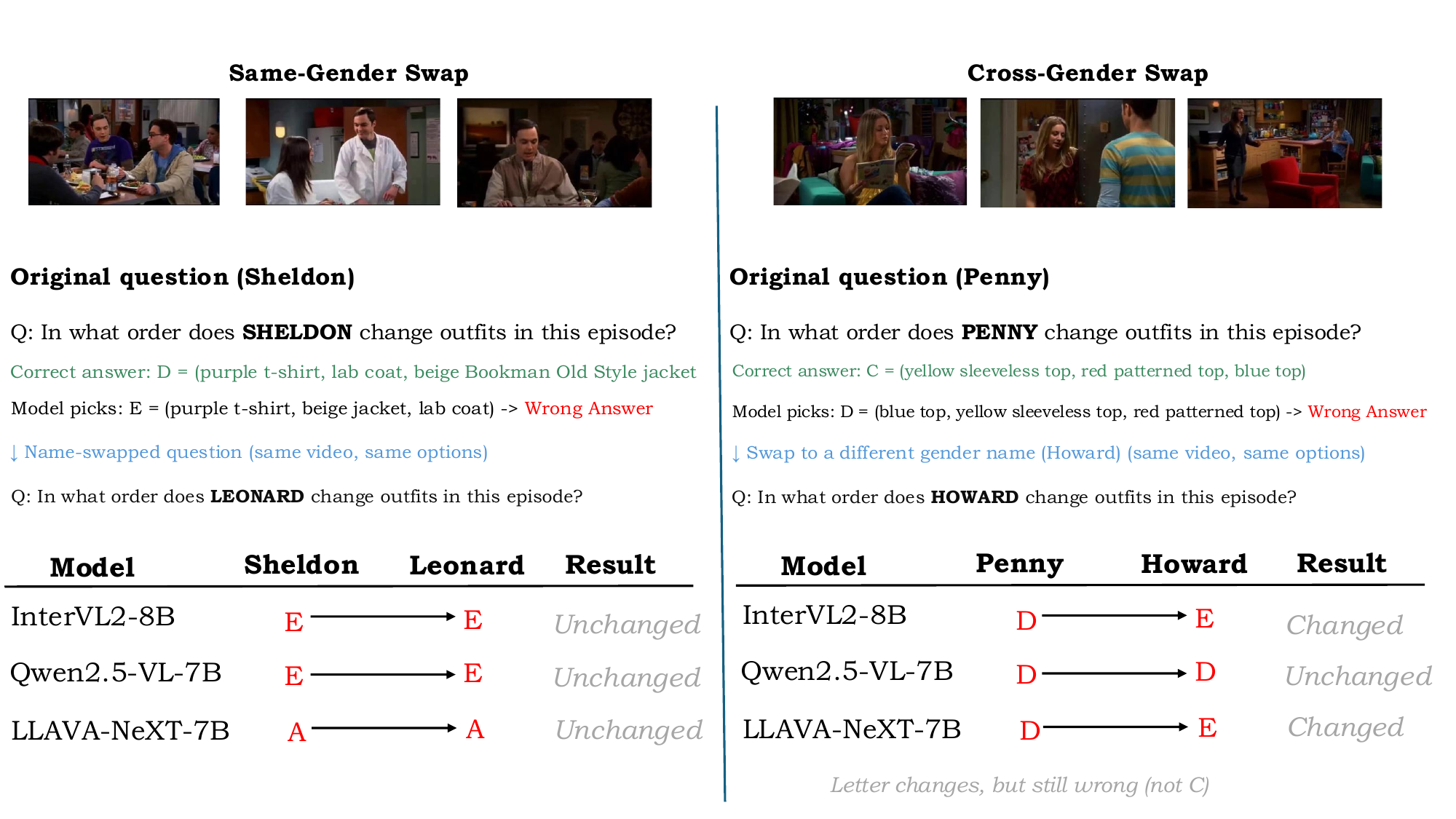}\par
  \vspace{0.2em}
  {\small\refstepcounter{figure}\label{fig:teaser}\noindent\textbf{Figure~\thefigure:} \textbf{Name substitution diagnostic.} \emph{Same-gender} (left): on Sheldon$\to$Leonard, all three models keep the same letter, they do not condition on the name. \emph{Cross-gender} (right): on Penny$\to$Howard, two of three change letter but land on a wrong answer, a coarse gender cue, not identification. \emph{Bottom}: sensitivity over $n{=}124$ BBT swap pairs by gender. \textbf{Takeaway: models respond to gender, not identity, same-gender swaps move the letter in only 7--17\% of cases vs.\ 20--43\% cross-gender.}}
\end{center}
\vspace{-1.0em}
\end{strip}

\begin{abstract}
\noindent Can a Video Large Language Model (Video-LLM) follow one person through a long video, keeping track of who they are well enough to report, in order, how their outfit changes across a full TV episode? Benchmarks increasingly score this kind of task, and the strongest open-source 7--8B models now reach 37--38\% on InfiniBench's \emph{global appearance} task, which asks exactly that. But does that score come from tracking the named character, or from something easier? We test this with a nine-condition diagnostic protocol applied to three architecturally distinct open-source Video-LLMs, with Gemini~2.5~Flash as a frontier reference, and find the accuracy does not come from character tracking. When we change the character named in the question to a different cast member, leaving the video and answer options untouched, the models change their answer only 4--31\% of the time, so they are largely ignoring who the question asks about. Breaking that test down by the gender of the swapped name shows why: the models react more when the name is changed to a different-gender character than to a same-gender one
(a 13--28 point gap), picking up coarse gender cues but unable to tell same-gender individuals apart. This shallow processing surfaces again when we drop the multiple-choice options and ask the same questions open-endedly: open-source accuracy drops 18--25 points, with none of 151 answers fully correct, versus a 12-point drop for Gemini. Further checks rule out the obvious innocent explanations, adding subtitles, using the most informative frames, or doubling the number of frames all leave character tracking unimproved, so the bottleneck is not how much video the model sees but how it ties that video to the person the question names. We release a diagnostic toolkit for auditing what such benchmark scores actually measure: \href{https://anonymous.4open.science/r/videollm-diagnostics-B17E/README.md}{link to our toolkit}.
\end{abstract}

\section{Introduction}\label{sec:intro}

Open-source 7--8B Video Large Language Models
(Video-LLMs) score 37--38\% accuracy in our evaluation on InfiniBench's \emph{global appearance} task~\citep{ataallah2024infinibench}, exceeding the previous generation of open-source long-form video specialists. On its face this is rapid progress. But a headline number is only as meaningful as the capability it tracks: answering this task correctly requires following a \emph{named} character across a full, 20--40 minute episode and identifying the order in which their outfit changes. Does a higher score mean the model is actually doing that?

There is reason for doubt. In image understanding, high benchmark accuracy has repeatedly masked shallow strategies: multimodal models that score well on chart, spatial-reasoning, and object-recognition benchmarks fail elementary questions about the very same inputs~\citep{razeghi2024plottwist, kamath2023whatsup, li2023pope}. Video adds another failure mode, many Video-LLMs are invariant to frame-shuffling~\citep{cores2024tvbench, xiao2024videoqa}, using little of the temporal structure their scores seem to require. Yet whether these models track character \emph{identity} at all, the prerequisite for answering ``what does Sheldon wear'' rather than ``what does someone wear'', has not been tested directly.

We test it. On InfiniBench's global appearance task, chosen because each question names a single character, letting us change only that name while holding the video and options fixed, we apply a nine-condition diagnostic protocol to three architecturally distinct open-source Video-LLMs (InternVL2-8B, Qwen2.5-VL-7B, LLaVA-NeXT-Video-7B), with Gemini~2.5~Flash as a frontier reference. The result is consistent across models (Figure~\ref{fig:teaser}): swapping the named character for another main-cast member changes the answer in only 4--31\% of cases, closer to ignoring the name than to tracking it. The accuracy gains are real; character tracking is not what produced them.

\paragraph{Contributions.} We contribute: \textbf{(1)} a nine-condition diagnostic protocol separating benchmark accuracy from character tracking; \textbf{(2)} an evaluation of three open-source Video-LLMs plus Gemini~2.5~Flash, including a gender-based decomposition of name substitution; and \textbf{(3)} a released toolkit for auditing character-tracking claims.

\section{Related Work}\label{sec:related}

\citet{ataallah2024infinibench} introduce InfiniBench, whose global appearance task asks for the chronological order of a named character's outfit changes across a full episode. We adopt this task but analyze the mechanisms behind its aggregate accuracy.

\paragraph{Diagnostic studies of multimodal models.} \citet{razeghi2024plottwist} use synthetic plus real-world chart questions to show that high ChartQA scores mask poor elementary chart understanding, they found that Gemini Pro Vision accuracy scores drops from 67\% on ChartQA to 58\% on their elementary questions, and collective accuracy (all questions about one figure correct) is below 36\% for the best model. \citet{wu2024chartinsights} find similar gaps between high-level and low-level chart QA. \citet{kamath2023whatsup} introduce position-prompt perturbations and find vision-language models cannot reliably distinguish ``object on the left'' from ``object on the right.'' \citet{li2023pope} show object hallucination undermines Visual Question Answering (VQA) accuracy\@. We adopt the spirit of these works for the video setting, with character names as our perturbation axis.

\paragraph{Temporal insensitivity in Video-LLMs.} \citet{cores2024tvbench} show that frame-shuffling barely affects accuracy on most temporal video QA benchmarks; \citet{xiao2024videoqa} report similar shuffling invariance for grounding-style questions. These findings establish that Video-LLMs frequently disregard frame order. We see character-identity sensitivity as an orthogonal axis on the same broader question: when a Video-LLM scores high on a benchmark, what information is it actually using? Frame shuffling tested temporal information; we test character-identity information. Both belong to the same diagnostic family.

\paragraph{Position bias.} \citet{zheng2024positionbias} document that LLMs systematically prefer specific answer letters in multiple choice settings independent of content. We confirm this effect generalizes to Video-LLMs and find different default letters per model, ruling out shared training artifacts as the sole cause.

\section{Experimental Setup}\label{sec:setup}

\subsection{Task and Data}

We use InfiniBench's \citep{ataallah2024infinibench} \textbf{global appearance} task, which presents an episode video and asks ``Choose the correct option for the following question: In what order does \{character\} change outfits in this episode?'' with five candidate orderings (A--E). Ground-truth orderings are derived from TVQA+ \citep{lei2020tvqaplus} character bounding boxes manually filtered to unique outfits per episode.

Episodes range from roughly 20 minutes (Big Bang Theory, Friends) to 40 minutes (Castle). The open-source models receives 16 frames sampled uniformly across the episode, from which it must infer a named character's full sequence of outfit changes. On the BBT validation split we use, each episode carries 5--8 global-appearance questions on average.

InfiniBench's test split has hidden labels (leaderboard-only) and cannot be used for local diagnostic analysis. We therefore evaluate primarily on the \textbf{validation split}, which contains 129 BBT (\textit{The Big Bang Theory}) questions for global appearance. The validation split is the standard target for diagnostic analysis of contemporary benchmarks: it has visible labels and is held out from typical pretraining-data leaks. For cross-show generalization analysis we additionally use the \textbf{train split} for Friends ($n=162$) and Castle ($n=196$), because those shows are not represented in the validation split. We verified zero overlap between the splits via question ID\@. Using train-split data is, if anything, a stricter test: a model that fails on data it may have seen in pretraining is more telling than one that fails on held-out data.

\subsection{Models}

We evaluate three open-source 7--8B Video-LLMs from architecturally distinct families plus Gemini 2.5 Flash as a frontier reference:

\begin{itemize}
\setlength\itemsep{1pt}
\item \textbf{InternVL2{-}8B} \citep{chen2024internvl2}: InternViT vision encoder + InternLM2 language model.
\item \textbf{Qwen2.5{-}VL{-}7B} \citep{wang2025qwen25vl}: Qwen-specific vision tower + Qwen2.5 LLM\@.
\item \textbf{LLaVA{-}NeXT{-}Video{-}7B} \citep{zhang2024llavanextvideo}: CLIP-based vision encoder + Vicuna LLM\@.
% FIX in custom.bib: this citation must point to a Gemini 2.5 source, NOT
% the Gemini 1.5 report. (Table 2's "Gemini 1.5 Flash" row is correct.)
\item \textbf{Gemini 2.5 Flash} \citep{geminiteam2024gemini}: Frontier proprietary model with native long-video processing.
\end{itemize}

\paragraph{Model selection rationale.} We study three open-source Video-LLMs from architecturally distinct families, InternVL2-8B, Qwen2.5-VL-7B, and {LLaVA-NeXT-Video-7B}, so that any failure mode common to all three cannot be attributed to a quirk of a single architecture. We add Gemini~2.5~Flash as a frontier reference point. These models are widely used and sit in the 7--8B open-source tier that is most commonly deployed in practice, making their behavior on this task broadly representative.

\paragraph{Gemini scope.} We evaluate Gemini 2.5 Flash on a partial set of conditions, full coverage was beyond our API budget, and we mark its absence in tables (see \S\ref{sec:limitations}).

\subsection{Inference, Statistics, and Scope}\label{sec:scope}

\paragraph{Inference configuration.} For each open-source model we use 16 uniformly sampled frames per episode (matching InfiniBench's reported open-source evaluation protocol) at the model's native resolution, greedy decoding (\texttt{temperature=0}), and \texttt{max\_new\_tokens=10} for multiple-choice and \texttt{max\_new\_tokens=512} for open-ended. Gemini 2.5 Flash processes the full video natively at provider-default settings, with \texttt{max\_output\_tokens=2048} to accommodate internal reasoning tokens.

\paragraph{Statistical methods.} We report Wilson 95\% confidence intervals \citep{wilson1927confidence} on every accuracy and sensitivity, McNemar's exact test \citep{mcnemar1947note} for paired comparisons, chi-squared tests of the predicted-letter distribution against uniform for position bias, and Cohen's $h$ \citep{cohen1988statistical} for effect sizes (0.2/0.5/0.8 = small/medium/large).

\paragraph{Experimental scope.} Sample sizes are stated inline and summarized in the supplement. We evaluate all four models on BBT validation, our canonical diagnostic target because it has visible labels; the three open-source models are also evaluated on Friends and Castle train splits where cross-show or annotation-window analyses require them. Gemini 2.5 Flash is evaluated on a partial subset, full-episode BBT, open-ended BBT, name substitution at $n=23$, collective accuracy, and question complexity, because a full diagnostic sweep exceeded our API budget (see \S\ref{sec:limitations}).

% [MOVED TO SUPPLEMENTARY: Table 1 "Experimental scope across all conditions"
%  (per-condition sample sizes for all four models). Reinsert as Table S1.]

\begin{table}[t]
\centering
\small
\caption{\textbf{Main results.} This table represents the older open-source baselines (italicized) that are reported by \citet{ataallah2024infinibench} on their full global appearance evaluation set spanning multiple shows; our numbers are on the BBT validation subset. These ranges are not strictly comparable but provide order-of-magnitude context for where current 7--8B models sit relative to earlier long-video systems. The brackets represents the Wilson 95\% CIs.}
\label{tab:main}%
\resizebox{\columnwidth}{!}{%
\begin{tabular}{lccccc}
\toprule
Model & $n$ & Full episode & Text-only & Pos.\ bias ($\chi^2$) & Most-pred letter \\
\midrule
\textit{Random} & --- & 20.0 & 20.0 & --- & --- \\
\textit{LLaMA-VID} & full eval$^{\ddagger}$ & 9.47 & --- & --- & --- \\
\textit{Goldfish} & full eval$^{\ddagger}$ & 9.62 & --- & --- & --- \\
\textit{Gemini 1.5 Flash} & full eval$^{\ddagger}$ & 33.31 & --- & --- & --- \\
\textit{GPT-4o} & full eval$^{\ddagger}$ & 46.84 & --- & --- & --- \\
\midrule
LLaVA{-}NeXT{-}Video{-}7B (ours)        & 129 BBT & 23.3 [16.7, 31.5] & 16.3 [10.9, 23.6] & --- & A (70\%) \\
InternVL2{-}8B (ours)               & 129 BBT & 38.0 [30.1, 46.6] & 18.6 [12.8, 26.2] & 58.64$^{***}$ & E (42\%) \\
Qwen2.5{-}VL{-}7B (ours)              & 129 BBT & 37.2 [29.4, 45.8] & 25.6 [18.8, 33.7] & 29.26$^{***}$ & E (36\%) \\
\midrule
Gemini 2.5 Flash (ours)           & 129 BBT & 61.2 [52.6, 69.2] & --- & --- & --- \\
\bottomrule
\end{tabular}%
}

\end{table}

\section{Diagnostic Protocol}\label{sec:protocol}

We define nine conditions; C1--C6 form the core protocol, and C7--C9 rule out alternative explanations for the failures the core protocol surfaces,namely too few frames, missing dialogue, and the inability to abstain.
\begin{itemize}
\setlength\itemsep{1pt}
\item \textbf{C1: Full-episode} standard InfiniBench Multiple Choice Questions on the full video.
\item \textbf{C2: Text-only} same prompt with 16 black frames, isolating accuracy obtainable from prompt and option text alone.
\item \textbf{C3: Targeted frames} --- replace uniform 16-frame sampling with oracle frames drawn directly from the ground-truth outfit-change windows (one frame per window). If accuracy does not rise even when the most informative frames are supplied, the relevant visual content is reaching the model, yet tracking does not improve.
\item \textbf{C4: Name substitution} replace the named character with a randomly chosen other main-cast character, leaving video and options identical; an identity-conditioned model should change its answer, whereas a position-bias-driven answer is unchanged by construction.
\item \textbf{C5: Open-ended generation} elicit the outfit sequence in free text, scored against ground truth on a $\{0,0.5,1\}$ scale by an LLM judge (Claude) \citep{zheng2023judging}; validated against a human rater in \S\ref{sec:openended}.
\item \textbf{C6: Metadata-only} remove the video, keep only the show title, to test pretraining contamination; near-chance accuracy rules it out.
\item \textbf{C7: Frame-count ablation} re-run at 8/16/32/64 frames on InternVL2 and Qwen; if accuracy rises but name sensitivity does not, more frames aid feature extraction, not tracking.
\item \textbf{C8: Abstention} add ``F. I cannot determine this from the video'' to all 129 BBT questions, testing whether models recognize when a question exceeds their capability.
\item \textbf{C9: Subtitle ablation} re-run with parsed dialogue text (capped at 4000 characters) to test whether missing textual context drives the failures.
\end{itemize}

\section{Results}\label{sec:results}

\subsection{Headline Accuracy and Position Bias}\label{sec:headline}

Table~\ref{tab:main} reports our main accuracy results alongside InfiniBench's reported numbers for older open-source systems. Three findings:

\paragraph{Newer 7--8B models substantially exceed prior open-source results.} We evaluate three current-generation 7--8B Video-LLMs whose families (InternVL, Qwen2.5-VL) are also represented in \citet{ataallah2024infinibench}'s recent evaluations on their full multi-show global appearance set. Our BBT-validation numbers are not directly comparable to theirs due to subset differences, but both works find that these newer 7--8B models exceed the accuracy of the earlier long-video specialists. Where prior work reports aggregate accuracy.

\paragraph{Position bias is highly significant for all open-source models.}
InternVL2 predicts answer letter E in 42\% of the 129 questions, despite E being the correct answer only 24\% of the time. We test whether this imbalance is consistent with random selection using a chi-squared test against a uniform distribution (the distribution we would expect if the model picked letters at random). For InternVL2, $\chi^2 = 58.64$ with $p = 5.6 \times 10^{-12}$: an effect this large is extraordinarily unlikely under the random-selection null hypothesis, well below the conventional $p < 0.05$ threshold.\footnote{We use $p < 0.05$ throughout as the conventional significance threshold; our reported $p$-values are typically many orders of magnitude smaller.} Qwen2.5-VL-7B shows the same effect with a different magnitude: it predicts E in 36\% of questions ($\chi^2 = 29.26$, $p = 6.9 \times 10^{-6}$). LLaVA-NeXT-Video-7B predicts A in 65\% of questions.

These distributions (42\% E, 36\% E, 65\% A) are statistically incompatible with the $\sim$20\%-per-letter pattern of random guessing, but ``not random'' does not mean ``informed'': the models pick by a learned per-letter bias, not by video content.

The fact that the three models prefer different default letters (E for InternVL2 and Qwen, A for LLaVA) argues against a single shared cause, such as common training data. The bias appears to be model-specific, and identifying its exact mechanism is beyond the scope of this paper. We leave a mechanistic investigation to future work.

\textbf{Gemini 2.5 Flash anchors the upper end.} Gemini 2.5 Flash achieves 61.2\% on our BBT evaluation, comparable in scale to GPT-4o's 46.84\% reported by InfiniBench on their full evaluation set. The two numbers are not strictly comparable (different evaluation subsets), but both indicate that frontier-tier models score substantially above the 23--38\% range observed for open-source 7--8B models. We use Gemini 2.5 Flash as a single frontier reference point and do not run the full diagnostic protocol on it; see \S\ref{sec:limitations} for discussion.

\subsection{Video Contributes Over Text-Only}
To test whether the video provides meaningful signal beyond what is available from the question and options alone, we compare InternVL2's accuracy on the same 129 BBT questions with and without video, using McNemar's paired test. On 31 of the 129 questions, the model gets the answer right when given the video and wrong when given black frames; on only 6 questions does removing the video flip an answer from wrong to right. This 31-vs-6 asymmetry yields $p < 10^{-4}$ and Cohen's $h = 0.44$ (medium-to-large effect), so the video clearly provides usable signal: InternVL2 gains 19.4 percentage points of accuracy when the visual modality is available (38.0\% full-episode vs.\ 18.6\% text-only).
However, the name-substitution analysis in \S\ref{sec:nameswap} will show that this visual signal is not being used to identify \emph{which} character the question is about. The video helps the model recognize what outfits are in the scene, but does not bind those outfits to the character named in the question. The visual information is processed, but not bound to character-identity tokens.

\subsection{Subtitles Do Not Bridge the Gap}\label{sec:subtitles}

One possible explanation for the failures we report is that models lack sufficient context to identify which character's outfits the question concerns. The video may show outfits, but without dialogue cues (``Penny, your shirt is nice'') the model may not know who is who. To test this, we re-run the full-episode condition with the episode's parsed dialogue text (from the corresponding \texttt{.srt} subtitle file) included in the prompt, capped at 4000 characters to fit context limits.

Adding subtitles helps no model: LLaVA $23.3\!\to\!20.9$, InternVL2 $38.0\!\to\!35.7$, Qwen $37.2\!\to\!36.4$ (changes of $-2.3$, $-2.3$, $-0.8$ Percentage Points; McNemar $p=0.63$, $1.00$, $0.63$). Adding dialogue does not bridge the gap between aggregate accuracy and genuine character tracking. This rules out an alternative explanation for our results, namely, that the models could solve the task if only they had textual context to disambiguate character identity. They cannot; the failure is in the visual-to-identity binding, not in the availability of identifying text.

\subsection{Targeted Frames Reveal Annotation-Structure Confound}\label{sec:targeted}

\begin{table}[ht]
\centering
\small
\caption{\textbf{Targeted-frames condition} (one oracle frame per ground-truth annotation window, replacing uniform 16-frame sampling). $\Delta$ is the change in accuracy from the standard full-episode condition (Table~\ref{tab:main}) to the targeted-frames condition for the same model and show.}
\label{tab:targeted}%
\begin{tabular}{lcccc}
\toprule
Model & Show & $n$ & Acc. & $\Delta$ \\
\midrule
\multirow{3}{*}{InternVL2}
& BBT & 129 & 34.1 & $-3.9$ \\
& Friends & 132 & 31.1 & $+1.5$ \\
& Castle & 146 & 50.0 & $+20.4$ \\
\midrule
\multirow{3}{*}{Qwen}
& BBT & 129 & 34.1 & $-3.1$ \\
& Friends & 132 & 42.4 & $+7.2$ \\
& Castle & 146 & 47.9 & $+19.3$ \\
\midrule
\multirow{3}{*}{LLaVA}
& BBT & 129 & 23.3 & $0.0$ \\
& Friends & 132 & 26.5 & $+1.2$ \\
& Castle & 146 & 30.1 & $+9.7$ \\
\bottomrule
\end{tabular}
\end{table}

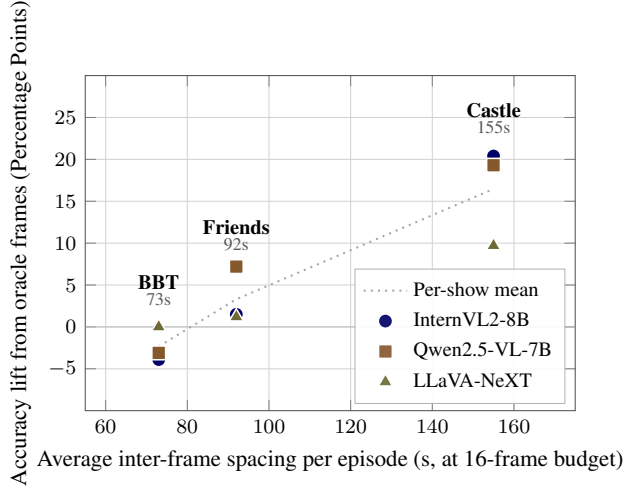
\begin{figure}[!t]
\centering
\resizebox{\columnwidth}{!}{%
\begin{tikzpicture}
  \begin{axis}[
      width=\linewidth,
      height=6.2cm,
      xlabel={Average inter-frame spacing per episode (s, at 16-frame budget)},
      ylabel={Accuracy lift from oracle frames (Percentage Points)},
      xlabel style={font=\small},
      ylabel style={font=\small},
      tick label style={font=\footnotesize},
      xmin=55, xmax=175,
      ymin=-10, ymax=30,
      xtick={60,80,100,120,140,160},
      ytick={-5,0,5,10,15,20,25},
      grid=both,
      grid style={line width=.2pt, draw=gray!25},
      major grid style={line width=.3pt, draw=gray!35},
      axis line style={draw=black!70},
      legend style={font=\footnotesize,
        draw=black!30,
        fill=white,
        fill opacity=0.92,
        text opacity=1,
        at={(0.98,0.02)},
        anchor=south east,
        legend columns=1,
        row sep=1pt,
        inner sep=3pt,
      },
      legend cell align={left},
      enlargelimits=false,
      clip=false,
  ]
  % Zero reference line
  \addplot[domain=55:175, samples=2, gray!50, thin, forget plot] {0};

  % Per-show mean trend (dotted, no marker)
  \addplot[dotted, gray!70, thick, mark=none]
      coordinates {(73, -2.33) (92, 3.30) (155, 16.47)};
  \addlegendentry{Per-show mean}

  % InternVL2-8B (slate blue circles)
  \addplot[only marks, mark=*, mark size=2.8pt,
           color=blue!40!black, fill=blue!40!black,
           mark options={draw=white, line width=0.6pt}]
      coordinates {(73, -3.9) (92, 1.5) (155, 20.4)};
  \addlegendentry{InternVL2-8B}

  % Qwen2.5-VL-7B (warm brown squares)
  \addplot[only marks, mark=square*, mark size=2.7pt,
           color=brown!70!black, fill=brown!70!black,
           mark options={draw=white, line width=0.6pt}]
      coordinates {(73, -3.1) (92, 7.2) (155, 19.3)};
  \addlegendentry{Qwen2.5-VL-7B}

  % LLaVA-NeXT-Video-7B (muted olive triangles)
  \addplot[only marks, mark=triangle*, mark size=3.2pt,
           color=olive!60!black, fill=olive!60!black,
           mark options={draw=white, line width=0.6pt}]
      coordinates {(73, 0.0) (92, 1.2) (155, 9.7)};
  \addlegendentry{LLaVA-NeXT}

  % Show labels positioned above each cluster, clearly out of the way
  \node[font=\footnotesize\bfseries, anchor=south]
       at (axis cs:73, 3.5) {BBT};
  \node[font=\scriptsize, anchor=south, gray!70!black]
       at (axis cs:73, 1.5) {73s};

  \node[font=\footnotesize\bfseries, anchor=south]
       at (axis cs:92, 10) {Friends};
  \node[font=\scriptsize, anchor=south, gray!70!black]
       at (axis cs:92, 8) {92s};

  \node[font=\footnotesize\bfseries, anchor=south]
       at (axis cs:155, 24) {Castle};
  \node[font=\scriptsize, anchor=south, gray!70!black]
       at (axis cs:155, 22) {155s};

  \end{axis}
\end{tikzpicture}%
}
\caption{\textbf{Annotation-structure confound across shows.}
Accuracy lift from oracle frame selection versus average spacing between uniformly sampled frames. Longer videos have wider frame spacing under the same 16-frame budget, so oracle frames help more when uniform sampling misses annotated outfit-change windows. Dotted line: per-show mean lift. \textbf{Takeaway: apparent per-show difficulty can shift by $\sim$20 Percentage Points because of frame coverage, not character-tracking capability.}}
\label{fig:annotation_confound}
\end{figure}

The targeted-frames condition (Table~\ref{tab:targeted}) reveals that the same model, evaluated under the same protocol, can show very different effective performance across shows for reasons that have nothing to do with the model's capability. We run this condition on InternVL2 and Qwen, our two strongest open-source models, across the three shows for which InfiniBench provides outfit-change annotation windows (BBT, Friends, Castle).

Replacing uniform 16-frame sampling with one oracle frame per ground-truth window does not help on BBT (McNemar $p=0.61$, Cohen's $h=0.06$), helps slightly on Friends ($+1.5$ Percentage Points InternVL2, $+7.2$ Percentage Points Qwen), and helps substantially on Castle ($\sim{+}20$ Percentage Points for both: 50.0\% vs.\ 29.6\% InternVL2; 47.9\% vs.\ 28.6\% Qwen).

The driver is the interaction between episode duration and the fixed 16-frame budget: uniform sampling spaces frames every $\sim$73s on BBT (19.4 min), $\sim$92s on Friends (24.6 min), and $\sim$155s on Castle (41.3 min). When spacing approaches the duration of an outfit-change window, uniform frames land in annotated windows by chance and targeting adds little (BBT); when spacing exceeds it, uniform frames miss the windows and targeting helps substantially (Castle), with Friends intermediate (Figure~\ref{fig:annotation_confound}). This is not that Castle is intrinsically harder, nor that models improve with targeting: the uniform protocol gives different temporal resolution per show, so per-show accuracy partly reflects a $\sim$20 Percentage Points structural artifact rather than capability. We give the benchmark-design implication in \S\ref{sec:discussion}.

\subsection{Name Substitution: Models Are Insensitive to Character Identity}
\label{sec:nameswap}

\begin{table}[ht]
\caption{\textbf{Name substitution sensitivity.} Percentage of questions where the model produced a different answer letter when the character name in the question was swapped to a randomly chosen other main-cast character; the video and options are unchanged. Brackets show Wilson 95\% confidence intervals.}
\label{tab:nameswap}
\centering
\scriptsize
\setlength{\tabcolsep}{3pt}
\resizebox{\columnwidth}{!}{%
\begin{tabular}{lccc}
\toprule
Model & BBT & Friends & Castle \\
      & ($n=124$) & ($n=124$) & ($n=124$) \\
\midrule
LLaVA-NeXT-Video-7B$^\dagger$ & 13.7 [8.7, 20.9]  & 4.8 [2.1, 10.7]   & 4.0 [1.6, 9.6] \\
InternVL2-8B                  & 26.6 [19.6, 35.0] & 12.1 [7.4, 19.1]  & 9.7 [5.6, 16.2] \\
Qwen2.5-VL-7B                 & 30.6 [23.2, 39.2] & 16.9 [11.2, 24.7] & 13.7 [8.7, 20.9] \\
\bottomrule
\end{tabular}%
}
\\[2pt]
{\footnotesize $^\dagger$LLaVA's baseline shows extreme A-bias (70\% of all answers are A);}\\
{\footnotesize cross-show sensitivity is correspondingly suppressed.}
\end{table}

For each question, we replace the named character with a randomly selected other main-cast character while keeping the video and options unchanged. Because different characters usually wear different outfits, a model that conditions on the named character should change its answer letter in most swapped cases. The swapped character's true outfit sequence need not appear among the original options A--E; the diagnostic only asks whether a name-conditioned model changes letter, whereas a model that ignores the name gives the same letter in both versions. Table~\ref{tab:nameswap} reports this sensitivity.

All three open-source models show name-substitution sensitivity well below 31\% on BBT and lower still on Friends and Castle (4--17\%). Even at the upper end of the Wilson 95\% confidence interval on BBT (35\% for InternVL2, 39\% for Qwen), the models change answer fewer than two times out of five; at $n=124$ per show, every Friends and Castle CI upper bound is below 25\%. Since a model that fully ignores the character name would have 0\% sensitivity, the observed 4--31\% range is much closer to ignoring the name than to genuine tracking. \emph{Models are not conditioning on character identity in any robust way.}

\subsubsection{Mechanism: Gender-Decomposed Sensitivity}
\label{sec:gender}

To understand which swaps moved the answer, we partition the same 124 BBT swap pairs from Table~\ref{tab:nameswap} into same-gender and cross-gender swaps. For the seven main BBT characters (Sheldon, Leonard, Howard, Raj male; Penny, Bernadette, Amy female), same-gender swaps are e.g.\ Sheldon$\to$Howard and cross-gender swaps are e.g.\ Sheldon$\to$Penny.
\begin{table}[ht]
\centering
\small
\setlength{\tabcolsep}{4pt}
\caption{\textbf{Gender-decomposed name substitution} on BBT (124 pairs total from Table~\ref{tab:nameswap}, partitioned into same-gender and cross-gender subsets). Sensitivity rates are partitioned by whether the swap target is the same gender as the original character or a different gender; $\Delta$ is cross-gender minus same-gender.}
\label{tab:gender}
\resizebox{\columnwidth}{!}{%
\begin{tabular}{lccc}
\toprule
Model & Same-gen. & Cross-gen. & $\Delta$ \\
\midrule
LLaVA-NeXT-Video-7B & 6.8  & 20.0 & $+13.2$ \\
InternVL2-8B        & 11.9 & 40.0 & $+28.1$ \\
Qwen2.5-VL-7B       & 16.9 & 43.1 & $+26.1$ \\
\bottomrule
\end{tabular}%
}
\end{table}
Table~\ref{tab:gender} reveals the mechanism: all three models are much more sensitive to cross-gender than same-gender swaps, with deltas of 13--28 Percentage Points. \emph{Models use coarse gender visual cues but cannot distinguish individuals within gender.} Sheldon and Howard, both male and physically similar, are essentially interchangeable; Sheldon and Penny are not. Pair-level rates make this concrete: the five highest-sensitivity BBT pairs are nearly all cross-gender (e.g., Penny$\to$Leonard 100\%, Bernadette$\to$Sheldon 100\%, Howard$\to$Penny 75\%), while the five lowest are nearly all same-gender (Leonard$\to$Howard 0\%, Sheldon$\to$Howard 0\%). These pair-level rates are noisy because sample sizes range from roughly 3 to 12 questions, but the overall ordering is robust.

Thus the aggregate 14--31\% BBT name-substitution sensitivity reflects gender as a category-level filter, not robust character identity: models can separate Sheldon from Penny, but not Sheldon from Howard, and otherwise guess.

\subsection{Open-Ended Evaluation Reveals Forced Guessing}
\label{sec:openended}

\begin{table}[ht]
\centering
\scriptsize
\setlength{\tabcolsep}{4pt}
\caption{\textbf{Multiple Choice (MC) vs.\ Open-Ended (OE)} on BBT ($n=47$--$52$). Effective accuracy is the mean $\{0, 0.5, 1\}$ score. Gap is OE effective accuracy minus MC.}
\label{tab:openended}
\begin{tabular}{lcccc}
\toprule
Model & MC & OE & Gap & Fully Correct \\
\midrule
LLaVA-NeXT-Video-7B & 23.3 & 2.9  & $-20.4$ & 0/52 \\
InternVL2-8B        & 38.0 & 13.0 & $-25.0$ & 0/47 \\
Qwen2.5-VL-7B       & 37.2 & 19.2 & $-18.0$ & 0/52 \\
Gemini 2.5 Flash    & 61.2 & 49.0 & $-12.2$ & 14/52 \\
\bottomrule
\end{tabular}
\end{table}

To test whether Multiple Choice (MC) accuracy reflects genuine knowledge or forced selection from a constrained option set, we re-pose the same questions in Open-Ended (OE) form: ``Describe each outfit in the order it appears.'' Responses are scored on a $\{0, 0.5, 1\}$ scale against ground-truth descriptions using Claude as an LLM judge~\citep{dubois2024alpacaeval,zheng2023judging}; a 20-item subset re-scored by an independent human rater gives quadratic-weighted Cohen's $\kappa=0.63$, with all disagreements between 0.5 and 0.0 and none involving the fully-correct (1.0) category.

The MC-OE gap holds across all four models and scales inversely with model capability (Table~\ref{tab:openended}). Open-source 7--8B models show 18--25 Percentage Point gaps: LLaVA-NeXT-Video falls from 23.3\% MC to 2.9\% OE, InternVL2 from 38.0\% to 13.0\%, and Qwen from 37.2\% to 19.2\%, with zero of 151 open-ended responses fully correct. Gemini 2.5 Flash shows the same qualitative gap but less extremely, dropping from 61.2\% MC to 49.0\% OE (a 12.2 Percentage Points gap), with 14 of 52 fully correct.

The error types sharpen the distinction. Open-source failures are dominated by fabrication of outfits absent from the ground truth (LLaVA 81\%, InternVL2 72\%; Qwen splits 31\% fabrication / 31\% explicit refusal). Gemini's errors are mostly partial matches (44\%) with only 13\% pure fabrication, consistent with a model attempting identification but bottlenecked by visual precision rather than by failing to bind evidence to character identity.

Thus MC accuracy partly reflects the constraint of the option set rather than the ability to identify and describe character appearance from video. Gemini's smaller gap shows that OE evaluation separates frontier from open-source models more sharply than aggregate MC accuracy. This extends \citet{razeghi2024plottwist}'s chart-finding pattern, adequate multiple-choice scores but poor elementary open-ended answers, to long-form video character tracking.

\subsection{Calibration: Models Cannot Identify Their Own Uncertainty}
\label{sec:idk}

The original InfiniBench global appearance questions already include an ``I don't know'' choice among the five options (A--E), present in all 129 BBT validation questions at varying positions, but models select it rarely (InternVL2 2.3\%, Qwen 8.5\%, LLaVA 4.7\%). We therefore add a sixth option, ``F.\ I cannot determine this from the video,'' to all 129 questions and re-evaluate. Abstention remains low, so the failure to abstain is not an artifact of wording.

InternVL2 abstains in only 3.1\% of questions, with conditional accuracy statistically indistinguishable from baseline, giving no usable calibration signal. Qwen abstains more often (11.6\%) and gains $+5.8$ Percentage Points in conditional accuracy when committing, suggesting weak partial calibration, but still answers 50.4\% of \emph{committed} questions wrong. LLaVA-NeXT-Video-7B never abstains with the IDK option (0/129), consistent with its extreme A-bias (70\% A-prediction) dominating regardless of frame budget or abstention option.

Thus aggregate MC accuracy cannot be improved by abstention thresholding for these models: they do not know what they do not know.

\subsection{Frame-Count Ablation}
\label{sec:frames}

\begin{table}[ht]
\centering
\small
\caption{\textbf{Frame-count vs.\ name sensitivity} for Qwen on BBT ($n=124$ pairs). More frames raise accuracy but not sensitivity.}
\label{tab:frames-nameswap}
\begin{tabular}{lcc}
\toprule
Condition & Accuracy & Name sensitivity \\
\midrule
Qwen 16 frames & 37.2 & 30.6 [23.2, 39.2] \\
Qwen 32 frames & 46.0 & 28.2 [21.1, 36.7] \\
$\Delta$ & $+8.8$ & $-2.4$ \\
\bottomrule
\end{tabular}
\end{table}
We run the frames-vs-sensitivity comparison on Qwen, which showed the largest accuracy gain from additional frames and is therefore the most informative test of whether accuracy gains translate into tracking gains. Across 8/16/32(/64) frames, InternVL2 is flat (38.0\%/38.0\%/41.1\%; 64f exceeds its 8192-token context), while Qwen rises then falls (32.6\%/37.2\%/46.5\%/41.9\%). LLaVA-NeXT-Video-7B is also roughly flat (19.4\%, 23.3\%, 21.7\%, 18.6\%), with no monotonic improvement and a slight drop at 64 frames, consistent with its extreme position bias (70\% A-prediction; see \S\ref{sec:results}) dominating regardless of frame budget.

The decisive comparison is Table~\ref{tab:frames-nameswap}: Qwen's 16-to-32-frame accuracy gain ($+8.8$ Percentage Points) is \emph{not} accompanied by improved name sensitivity ($-2.4$ Percentage Points, within overlapping 95\% confidence intervals). Doubling the frame budget improves visual feature extraction, making the model a better outfit recognizer in general, but not better at identifying \emph{which character's} outfits the question concerns. Thus name-substitution insensitivity is not a frame-sampling bottleneck; the bottleneck is binding visual information to character identity, not how much video is available.

\subsection{Collective Accuracy: Episode-Level Comprehension}
\label{sec:collective}

Aggregate accuracy can hide scattered correctness without episode-level comprehension. Following \citet{razeghi2024plottwist}, we compute \emph{collective accuracy}: the fraction of episodes where every question is answered correctly. The 129 BBT validation questions span 20 episodes (5--8 each); we use BBT to avoid potentially contaminated train-split data.

\begin{table}[ht]
\centering
\small
\caption{\textbf{Per-episode collective accuracy} on BBT ($n=20$ unique episodes covering the 129 validation questions). ``All correct'' counts episodes where the model answered every question about that episode correctly.}
\label{tab:collective}
\resizebox{\columnwidth}{!}{%
\begin{tabular}{lccc}
\toprule
Model & Aggregate & All correct & Rate \\
\midrule
LLaVA-NeXT-Video & 23.3 & 0/20 & 0.0\% \\
InternVL2-8B & 38.0 & 0/20 & 0.0\% \\
Qwen2.5-VL-7B & 37.2 & 0/20 & 0.0\% \\
Gemini 2.5 Flash & 61.2 & 1/20 & 5.0\% \\
\bottomrule
\end{tabular}%
}
\end{table}

Table~\ref{tab:collective} shows the gap between aggregate accuracy and per-episode comprehension. InternVL2-8B and Qwen2.5-VL-7B score 37--38\% in aggregate, yet neither gets \emph{all} questions correct on \emph{any} of the 20 BBT validation episodes. For InternVL2, 0/20 episodes are fully correct, 18/20 are partial, and 2/20 have no correct questions; the same pattern holds for the other open-source models. Even Gemini 2.5 Flash fully comprehends only 1/20 episodes (5\%). Thus 37--38\% aggregate accuracy means scattered correct answers, not understanding 37--38\% of episodes.

% [MOVED TO SUPPLEMENTARY: Sec "Question Complexity Breakdown" + Table 12
%  (accuracy by number of distinct outfits, 1/2/3/4+, for all four models).
%  Per-bucket point estimates are noisy (21--44 questions each); the
%  qualitative pattern is summarized in the Discussion. Reinsert as Sec S2.]

\subsection{Metadata-Only Baseline}
\label{sec:metadata}

One concern is contamination: because these shows are popular, a model may answer from pretraining memory rather than video, inflating accuracy without visual understanding. We test this with a metadata-only baseline: the video is removed and the prompt contains only the show name, question, and options. If contamination drove accuracy, this baseline should score well above chance. It does not: InternVL2 scores 17.8\% on BBT and 19.4\% on Castle, both at or below the 20\% chance level for five-way multiple choice. This rules out memorization of show titles, character names, or canonical outfits as a substantial driver of the 38\% full-episode accuracy we observe.

% [DELETED: Sec "Summary of Findings" (six-bullet dissociation recap).
%  Duplicated by the abstract and conclusion.]

\section{Discussion}
\label{sec:discussion}

\paragraph{What our findings imply about ``Video-LLMs.''} The four failure modes form one coherent picture. Consider a model that detects coarse visual categories (gender, clothing style) but never binds them to the named character. It would shift its answer on cross-gender swaps but not same-gender ones; score on MC when coarse cues narrow the options but fail open-ended generation that needs identity-bound descriptions; and have no signal separating ``I know'' from ``I don't,'' having never bound evidence to the name. All three open-source models match this profile.

One specific possibility worth noting is that BBT cast members are easily recognizable from the open internet, so models could in principle use pretraining-acquired face recognition or canonical-outfit associations to do well on this task. The combination of failure modes we identify argues against this being a major factor for the open-source models we evaluate: if the models could identify Sheldon by face, they would not confuse him with Howard under name swap, and the MC-OE gap would not be so large. We leave a direct test of pretrained face recognition's role in Video-LLM character tracking to future work.
\paragraph{Relationship to InfiniBench and prior diagnostic work.} Our work builds on \citet{ataallah2024infinibench}: their open-source numbers agree the task is hard, and the newer models simply score higher on the surface metric while failing the deeper diagnostics. The structure mirrors \citet{razeghi2024plottwist} on charts, high ChartQA accuracy masking poor elementary understanding, transposed to long-form video: 37--38\% MC accuracy overstates actual character-tracking ability.

\paragraph{Implications for benchmark design.} We draw four implications from this study:
\begin{itemize}
\setlength\itemsep{1pt}
\item Report name-substitution invariance alongside accuracy for named-character tasks: the two dissociate by 7--11 percentage points on BBT (more cross-show), so 38\% MC with 30\% sensitivity is not the advertised task.
\item Release per-episode annotation-window statistics (count, duration, dispersion) for fixed uniform-frame protocols: per-show difficulty varies by $\sim$20 Percentage Points when evenly spaced frames miss annotated outfit-change moments, not because tracking is worse.
\item Pair multiple-choice questions with an open-ended variant, or report the MC-OE gap: OE drops accuracy 18--25 points for open-source models (12 even for Gemini), with 0/151 open-source answers fully correct, showing MC partly reflects option-set constraint.
\item Balance correct-answer positions and report predicted-letter distributions: open-source models default to specific letters regardless of content (e.g., InternVL2 picks ``E'' 42\%, $\chi^2{=}58.64$), so scores can reflect letter-bias/key alignment rather than understanding.
\end{itemize}

\paragraph{Implications for model development.} The decoupling of accuracy gains from character-tracking capability ({\S\ref{sec:frames}}) suggests that current improvements to Video-LLMs are not improving tracking. Models scale better visual features, more reasoning, longer contexts, without addressing the binding of visual identity to language tokens. Our diagnostic protocol provides a direct way to test whether a model improvement specifically improves character tracking, distinct from improving aggregate accuracy.

\section{Limitations}
\label{sec:limitations}

\textbf{Sample size and granularity.} The main BBT analyses use $n=129$ questions, and open-source name substitution uses $n=124$ swap pairs per show for BBT, Friends, and Castle. These sizes support the large effects we report but limit finer subgroup analyses beyond coarse gender categories, such as sensitivity by actor, visual similarity, scene type, or outfit category.

\textbf{Video domain.} Our videos are scripted television, especially sitcom-style episodes with recurring characters, professional lighting, multi-camera editing, and distinctive wardrobes. This controlled setting may not map perfectly to messy real-world video, such as CCTV, body-camera footage, egocentric video, or crowded meetings with motion blur, occlusion, and unstable viewpoints.

\textbf{Model scope.} We focus on the current 7--8B open-source tier. Gemini~2.5~Flash is only a partial frontier reference, full-episode BBT, open-ended BBT, collective accuracy, question complexity, and 23 valid BBT name-swap pairs before API budget exhaustion, and we also cannot rule out different behavior in larger open-weight models.

\section{Conclusion}
\label{sec:conclusion}

Newer open-source Video-LLMs report higher long-form character-tracking accuracy than the previous generation, but our diagnostics show this accuracy does not reflect genuine character tracking. Across three architecturally distinct 7--8B models, name-substitution sensitivity stays below 31\% and is mostly limited to coarse cross-gender contrasts; open-ended evaluation drops multiple-choice accuracy by 18--25 percentage points, with 0/151 fully-correct open-source generations; collective accuracy is 0/20 BBT episodes; more frames improve accuracy without improving character identification; and ``cannot determine'' is almost never used. Together, these failures show that models use gender-level visual features as a category-level filter without binding visual identity to the character-name token. We extend InfiniBench by characterizing how and why current open-source models fail, and release our diagnostic protocol as a reusable toolkit for diagnosing character-tracking ability behind Video-LLM benchmark scores.

\begingroup
\small
\bibliographystyle{ieeenat_fullname}
\bibliography{custom}
\endgroup

\end{document}